\documentclass[11pt]{article}
\usepackage{acl2016}
\usepackage{times}
\usepackage{url}
\usepackage{latexsym}
\usepackage{graphicx}

\aclfinalcopy 

\title{Towards cross-lingual distributed representations without parallel text trained with adversarial autoencoders}

\author{Antonio Valerio Miceli Barone \\
  The University of Edinburgh \\
  Informatics Forum, 10 Crichton Street \\
  Edinburgh \\
  {\tt amiceli@inf.ed.ac.uk} \\}

\date{}

\begin{document}
\maketitle
\begin{abstract}
Current approaches to learning vector representations of text that are compatible between different languages usually require some amount of parallel text, aligned at word, sentence or at least document level. We hypothesize however, that different natural languages share enough semantic structure that it should be possible, in principle, to learn compatible vector representations just by analyzing the monolingual distribution of words.

In order to evaluate this hypothesis, we propose a scheme to map word vectors trained on a source language to vectors semantically compatible with word vectors trained on a target language using an adversarial autoencoder.

We present preliminary qualitative results and discuss possible future developments of this technique, such as applications to cross-lingual sentence representations.
\end{abstract}

\section{Introduction}
\label{SEC:INTRO}

Distributed representations that map words, sentences, paragraphs or documents to vectors real numbers have proven extremely useful for a variety of natural language processing tasks \cite{bengio2006neural,collobert2008unified,turian2010word,maas2011learning,mikolov2013distributed,socher2013parsing,pennington2014glove,levy2014neural,le2014distributed,baroni2014don,levy2015improving}, as they provide an effective way to inject into machine learning models general prior knowledge about language automatically obtained from inexpensive unannotated corpora. Based on the assumption that different languages share a similar semantic structure, various approaches succeeded to obtain distributed representations that are compatible across multiple languages, either by learning mappings between different embedding spaces \cite{mikolov2013exploiting,faruqui2014improving} or by jointly training cross-lingual representations \cite{klementiev-et-al:COLING2012,hermann2013multilingual,lauly2014autoencoder,gouws2014bilbowa}. These approaches all require some amount of parallel text, aligned at word level, sentence level or at least document level, or some other kind of parallel resources such as dictionaries \cite{ammar2016massively}.

In this work we explore whether the assumption of a shared semantic structure between languages is strong enough that it allows to induce compatible distributed representations without using any parallel resource. We only require monolingual corpora that are thematically similar between languages in a general sense.

We hypothesize there exist a suitable vectorial space such that each language can be viewed as a random process that produces vectors at some level of granularity (words, sentences, paragraphs, documents) which are then encoded as discrete surface forms, and we hypothesize that, if languages are used to convey thematically similar information in similar contexts, these random processes should be approximately isomorphic between languages, and that this isomorphism can be learned from the statistics of the realizations of these processes, the monolingual corpora, in principle without any form of explicit alignment.

We motivate this hypothesis by observing that humans, especially young children, who acquire multiple languages, can often do so with relatively little exposure to explicitly aligned parallel linguistic information, at best they may have access to distant and noisy alignment information in the form of multisensorial environmental clues. Nevertheless, multilingual speakers are always automatically able to translate between all the languages that they can speak, which suggests that their brain either uses a shared conceptual representations for the different surface features of each language, or uses distinct but near-isomorphic representations that can be easily transformed into each other.

\section{Learning word embedding cross-lingual mappings with adversarial autoencoders}
\label{SEC:MODEL}

The problem of learning transformations between probability distributions of real vectors has been studied in the context of generative neural network models, with approaches such as Generative Moment Matching Networks (GMMNs) \cite{li2015generative} and Generative Adversarial Networks (GANs) \cite{goodfellow2014generative}. In this work we consider GANs, since their effectiveness has been demonstrated in the literature more thoroughly than GMMNs.

In a typical GAN, we wish to train a \textit{generator} model, usually a neural network, to transform samples from a known, easy to sample, uninformative distribution (e.g. Gaussian or uniform) into samples distributed according to a target distribution defined implicitly by a training set. In order to do so, we iteratively alternate between training a differentiable \textit{discriminator} model, also a neural network, to distinguish between training samples and artificial samples produced by the generator, and training the generator to fool the discriminator into misclassifying the artificial examples as training examples. This can be done with conventional gradient-based optimization because the discriminator is differentiable thus it can backpropagate gradients into the generator.

It can be proven that, with sufficient model capacity and optimization power, sufficient entropy (information dimension) of the generator input distribution, and in the limit of infinite training set size, the generator learns to produce samples from the correct distribution. Intuitively, if there is any computable test that allows to distinguish the artificial samples from the training samples with better than random guessing probability, then a sufficiently powerful discriminator will eventually learn to exploit it and then a sufficiently powerful generator will eventually learn to counter it, until the generator output distribution becomes undistinguishable from the true training distribution. In practice, actual models have finite capacity and gradient-based optimization algorithms can become unstable or stuck when applied to this multi-objective optimization problem, though they have been successfully used to generate fairly realistic-looking images \cite{denton2015deep,radford2015unsupervised}.

\begin{figure}
\includegraphics[scale=0.5]{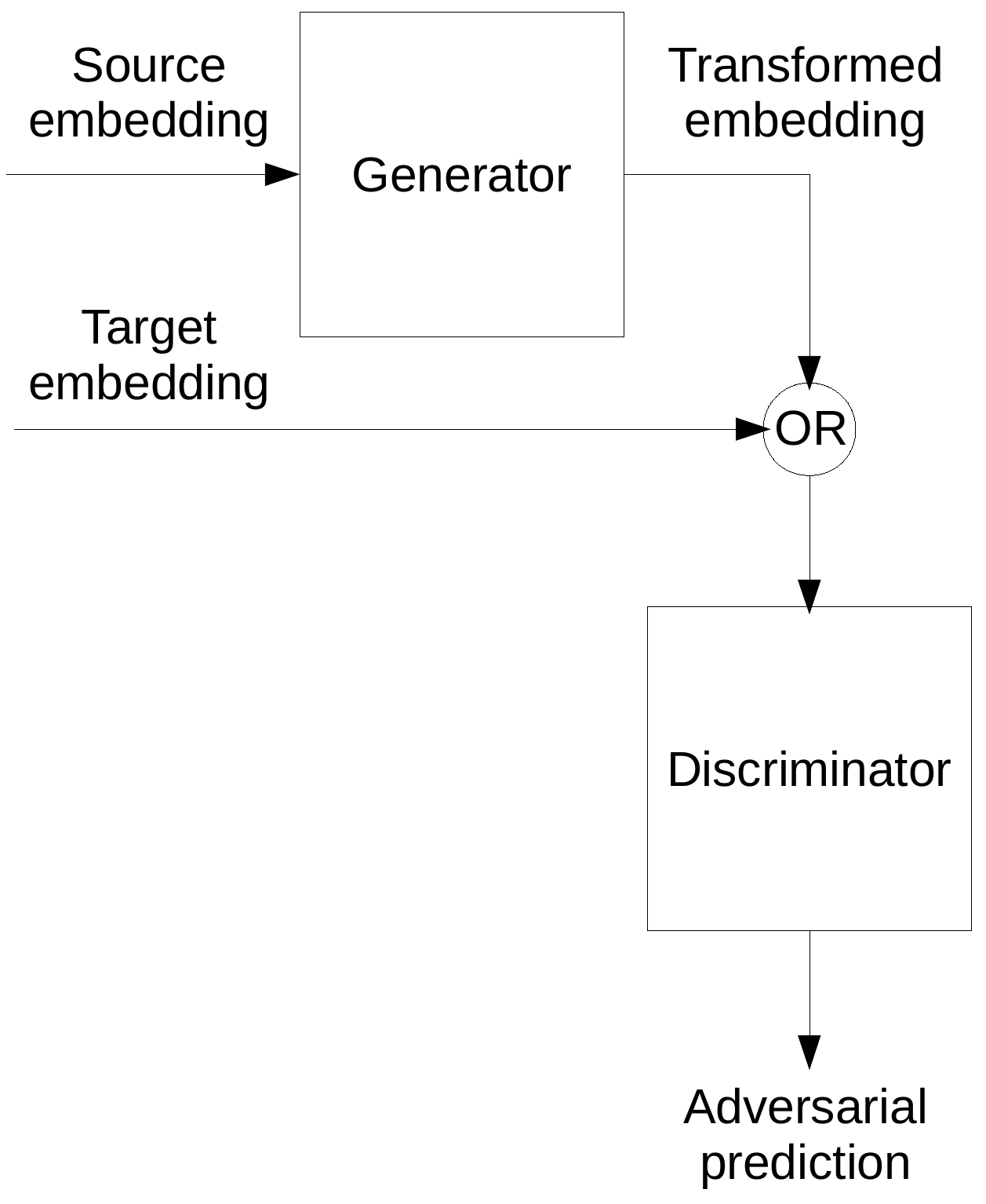} 
\caption{Generative adversarial network for cross-lingual embedding mapping}
\label{FIG:GAN}
\end{figure}

In our preliminary experiments we attempted to adapt GANs to our problem, by training the generator to learn a transformation between word embeddings trained on different languages (fig. \ref{FIG:GAN}). Let $d$ be the embedding dimensionality, $G_{\theta_G} \,:\, \mathcal{R}^d \rightarrow \mathcal{R}^d$ be the generator parametrized by $\theta_G$, $D_{\theta_D} \,:\, \mathcal{R}^d \rightarrow [0,1]$ be the discriminator parametrized by $\theta_D$.

At each training step:
\begin{enumerate}
\item draw a sample $\{f\}_n$ of $n$ source embeddings, according to their (adjusted) word frequencies 
\item transform them into target-like embeddings $\{\hat{e}\}_n = G_{\theta_G}(\{f\}_n)$
\item evaluate them with the discriminator, estimating their probability of having been sampled from the true target distribution $\{p\}_n = D_{\theta_D}(\{\hat{e}\})$
\item update the generator parameters $\theta_G$ to reduce the average adversarial loss $L_{a} = -\log(\{p\}_n)$
\item draw a sample $\{e\}_n$ of $n$ true target embeddings
\item update the discriminator parameters $\theta_D$ to reduce its binary cross-entropy loss on the classification between $\{e\}_n$ (positive class) and $\{\hat{e}\}$ (negative class)
\end{enumerate}
repeat these steps until convergence.

Unfortunately we found that in this setup, even with different network architectures and hyperparameters, the model quickly converges to a pathological solution where the generator always emits constant or near-constant samples that somehow can fool the discriminator. This appears to be an extreme case of the know mode-seeking issue of GANs \cite{radford2015unsupervised,theis2015note,salimans2016imprgan}, which is probably exacerbated in our settings because of the point-mass nature of our probability distributions where each word embedding is a mode on its own.

\begin{figure}
\includegraphics[scale=0.4]{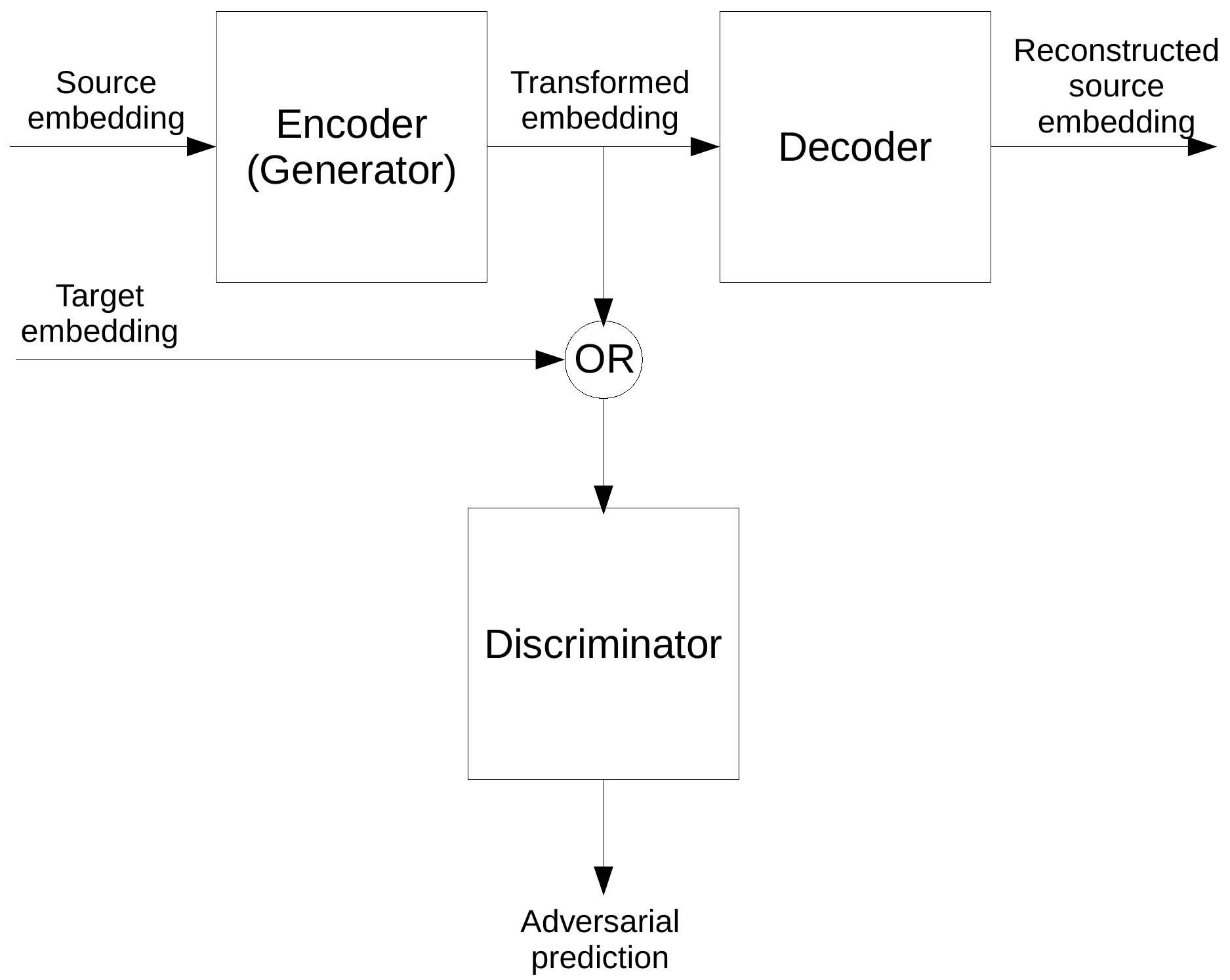} 
\caption{Adversarial autoencoder for cross-lingual embedding mapping (loss function blocks not shown).}
\label{FIG:AAE}
\end{figure}

In order to avoid these pathological solutions, we needed a way to penalize the generator for destroying too much information about its input. Therefore we turned our attention to Adversarial Autoencoders (AAE) \cite{makhzani2015adversarial}. In an AAE, the generator, now called \textit{encoder}, is paired with another model, the \textit{decoder} $R_{\theta_R} \,:\, \mathcal{R}^d \rightarrow \mathcal{R}^d$ parametrized by $\theta_R$ which attempts to transform the artificial samples emitted by the encoder back into the input samples. The encoder and the decoder are jointly trained to minimize a combination of the average reconstruction loss $L_{r}(\{f\}_n, R_{\theta_R}(G_{\theta_G}(\{f\}_n)))$ and the adversarial loss defined as above. The discriminator is trained as above. In the original formulation of the AAE, the discriminator is used to enforce a known prior (e.g. Gaussian or Gaussian mixture) on the intermediate, \textit{latent} representation, in our setting instead we use it to match the latent representation to the target embedding distribution so that the encoder can be used to transform source embeddings into target ones (fig. \ref{FIG:AAE}).

In our experiments, we use the cosine dissimilarity as reconstruction loss, and as a further penalty we also include the pairwise cosine dissimilarity between the generated latent samples $\{\hat{e}\}$ and the true target samples $\{e\}_n$. Therefore, the total loss incurred by the encoder-decoder at each step is

$L_{GR} =  \lambda_r L_{r}(\{f\}_n, R_{\theta_R}(G_{\theta_G}(\{f\}_n))) -\lambda_a \log(\{p\}_n) + \lambda_c L_{r}(\{e\}_n, G_{\theta_G}(\{f\}_n))$

where $\lambda_r, \lambda_a$ and $\lambda_c$ are hyperparameters (all set equal to 1 in our experiments).

\section{Experiments}
\label{SEC:EXPER}

We performed some preliminary exploratory experiments on our model. In this section we report salient results.

The first experiment is qualitative, to assess whether our model is able to learn any semantically sensible transformation at all. We consider English to Italian embedding mapping.

We train English and Italian word embeddings on randomly subsampled Wikipedia corpora consisting of about 1.5 million sentences per language. We use word2vec \cite{mikolov2013distributed} in skipgram mode to generate embeddings with dimension $d=100$. Our encoder and decoder are linear models with tied matrices (one the transpose of the other), initialized as random orthogonal matrices (we also explored deep non-linear autoencoders but we found that they make the optimization more difficult without providing apparent benefits).

Our discriminator is a Residual Network \cite{he2015deep} without convolutions, one leaky ReLU non-linearity \cite{maas2013rectifier} per block, no non-linearities on the passthrough path, batch normalization \cite{ioffe2015batch} and dropout \cite{srivastava2014dropout}. The block (layer) equation is:
\begin{equation}
h_{t+1} = \phi(W_t \times h_{t-1}) + h_{t-1}
\end{equation}  
 where $W_t$ is a weight matrix and $\phi$ is batch normalization (with its internal parameters) followed by leaky ReLU and $h_t$ is a $k$-dimensional block state (in our experiments $k=40$). The network has $T=10$ blocks followed by a $1$-dimensional output layer with logistic sigmoid activation. We found that using a Residual Network as discriminator rather than a standard multi-layer perceptron yields larger gradients being backpropagated to the generator, facilitating training. We actually train two discriminators per experiment, with identical structure but different random initializations, and use one to train the generator and the other for monitoring in order to help us determine whether overfitting or underfitting occurs.

At each step, word embeddings are sampled according to their frequency in the original corpora, adjusted to subsample frequent words, as in word2vec. Updates are performed using the Adam optimizer \cite{kingma2014adam} with learning rate $0.001$ for the encoder-decoder and $0.01$ for the discriminator.

The code\footnote{Code with full hyperparameters available at: https://github.com/Avmb/clweadv} is implemented in Python, Theano \cite{2016arXiv160502688short} and Lasagne.

We qualitatively analyzed the quality of the embeddings by considering the closest Italian embeddings to a sample of transformed English embeddings. We notice that in some cases the closest or nearly closest embedding is the true translation, for instance 'computer' (en) -\textgreater 'computer' (it). In other cases, the closest terms are not translations but subjectively appear to be semantically related, for instance 'rain' (en) -\textgreater  'gelo', 'gela', 'intensissimo', 'galleggiava', 'rigidissimo', 'arida', 'semi-desertico', 'fortunale', 'gelata', 'piovosa' (it 10-best),  or 'comics' (en) -\textgreater 'Kadath', 'Microciccio','Cugel','Promethea','flashback','episodio', 'Morimura', 'Chatwin', 'romanzato','Deedlit' (it 10-best), or 'anime' (en) -\textgreater 'Zatanna', 'Alita', 'Yuriko', 'Wildfire', 'Carmilla', 'Batwoman', 'Leery', 'Aquarion', 'Vampirella', 'Minaccia' (it 10-best). Other terms, such as names of places however, tend to be transformed incorrectly, for instance 'France' (en) -\textgreater 'Radiomobile', 'Cartubi', 'Freniatria', 'UNUCI', 'Cornhole', 'Internazione', 'CSCE', 'Folklorica', 'UECI', 'Rientro' (it 10-best).

We further evaluate our model on German to English and English to German embedding transformations, using the same evaluation setup as \cite{klementiev-et-al:COLING2012} with embeddings trained on the concatenation of the Reuters corpora and the News Commentary 2015 corpora, with embedding dimension $d=40$ and discriminator depth $T=4$. On a qualitative analysis notice similar partial semantic similarity patterns. However the cross-lingual document classification task we were able to improve over the baseline only for the smallest training set size.

\section{Discussion and future work}
\label{SEC:DISCUSS}

From the qualitative analysis of the word embedding mappings it appears that the model does learn to transfer some semantic information, although it's not competitive with other cross-lingual representation approaches. This may be possibly an issue of hyperparameter choice and architectural details, since, to our knowledge, this is the first work to apply adversarial training techniques to point-mass distribution arising from NLP tasks.

Further experimentation is needed to determine whether the model can be improved or whether we already hit a fundamental limit on how much semantic transfer can be performed by monolingual distribution matching alone. This additional experimentation may help to test how strongly our initial hypothesis of semantic isomorphism between languages holds, in particular across languages of different linguistic families.

Even if this hypothesis does not hold in a strong sense and semantic transfer by monolingual text alone turns out to be infeasible, our technique might help in conjunction with training on parallel data. For instance, in neural machine translation "sequence2sequence" transducers without attention \cite{cho2014properties}, it could be useful to train as usual on parallel sentences and train in autoencoder mode on monolingual sentences, using an adversarial loss computed by a discriminator on the intermediate latent representations to push them to be isomorphic between languages. A modification of this technique that allows for the latent representation to be variable-sized could be also applied to the attentive "sequence2sequence" transducers \cite{bahdanau2014neural}, as an alternative or in addition to monolingual dataset augmentation by backtranslation \cite{sennrich2015improving}.

Furthermore, it may be worth to evaluate additional distribution learning approaches such as the aforementioned GMMs, as well as the more recent BiGAN/ALI framework \cite{donahue2016adversarial,dumoulin2016adversarially} which uses an adversarial discriminator loss both to match latent distributions and to enforce reconstruction, and also to consider more recent GAN training techniques \cite{salimans2016imprgan}.

In conclusion we believe that this work initiates a potentially promising line of research in natural language processing consisting of applying distribution matching techniques such as adversarial training to learn isomorphisms between languages.

\section*{Acknowledgements}
\label{SEC:ACK}

This project has received funding from the European Union's Horizon 2020 research and innovation programme under grant agreement 645452 (QT21).

\bibliography{acl2016}
\bibliographystyle{acl2016}

\appendix

\end{document}